\documentclass[sigconf]{acmart}
\usepackage{algorithm}
\usepackage{algorithmic}
\usepackage{multicol}
\usepackage{multirow}
\usepackage{subfigure} 

\AtBeginDocument{%
  \providecommand\BibTeX{{%
    \normalfont B\kern-0.5em{\scshape i\kern-0.25em b}\kern-0.8em\TeX}}}

\setcopyright{acmcopyright}
\copyrightyear{2023}
\acmYear{2023}
\acmDOI{XXXXXXX.XXXXXXX}

\acmConference[WWW '23]{. In Proceedings of the Web Conference 2023}{April 30 - May 4, 2023}{TX, US}
\author{Yuting Yang}
  \affiliation{
  \institution{Key Lab of Intelligent Information Processing, Institute of Computing Technology,
Chinese Academy of Sciences; University of Chinese Academy of Sciences}
  \city{Beijing}
  \country{China}}
  \email{yangyuting@ict.ac.cn}

\author{Wenqiang Lei}
  \affiliation{
  \institution{Sichuan University}
  \city{Sichuan}
  \country{China}}
  \email{wenqianglei@gmail.com}

\author{Pei Huang}
  \affiliation{
  \institution{Stanford University}
  \city{California}
  \country{USA}}
  \email{huangpei@stanford.edu}

  \author{Juan Cao}
  \affiliation{
  \institution{Key Lab of Intelligent Information Processing, Institute of Computing Technology,
Chinese Academy of Sciences; University of Chinese Academy of Sciences}
  \city{Beijing}
  \country{China}}
  \email{caojuan@ict.ac.cn}

\author{Jintao Li}
  \affiliation{
  \institution{Key Lab of Intelligent Information Processing, Institute of Computing Technology,
Chinese Academy of Sciences}
  \city{Beijing}
  \country{China}}
  \email{jtli@ict.ac.cn}

\author{Tat-Seng Chua}
  \affiliation{
  \institution{National University of Singapore}
  \country{Singapore}}
  \email{dcscts@nus.edu.sg}
  
\begin{document}

\title{A Dual Prompt Learning Framework for Few-Shot Dialogue State Tracking}

\begin{abstract}
Dialogue state tracking (DST) module is an important component for task-oriented dialog systems to understand users' goals and needs. Collecting dialogue state labels including slots and values can be costly, especially with the wide application of dialogue systems in more and more new-rising domains. 
In this paper, we focus on how to utilize the language understanding and generation ability of pre-trained language models for DST. We design a dual prompt learning framework for few-shot DST. Specifically, we consider the learning of slot generation and value generation as dual tasks, and two prompts are designed based on such a dual structure to incorporate task-related knowledge of these two tasks respectively. In this way, the DST task can be formulated as a language modeling task efficiently under few-shot settings. Experimental results on two task-oriented dialogue datasets show that the proposed method not only outperforms existing state-of-the-art few-shot methods, but also can generate unseen slots. It indicates that DST-related knowledge can be probed from PLM and utilized to address low-resource DST efficiently with the help of prompt learning.
\end{abstract}



\maketitle

\section{Introduction}

\begin{figure}
\centering
\includegraphics[width=0.5\textwidth]{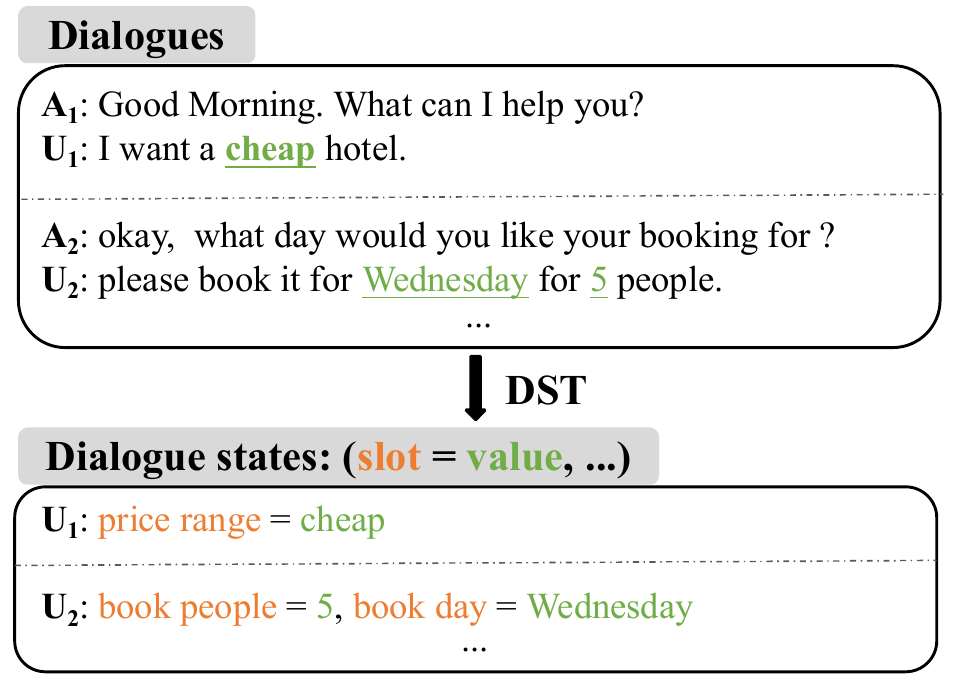}
\caption{Dialogue state tracking (DST) task. U and A represent the user's and system's utterances respectively. DST aims to extract dialogue states pairs (\textcolor{orange}{slot}, {\color{OliveGreen}{value}}), for each user's utterance. Values are usually the explicit needs expressed in the utterances.}
\label{DST}
\end{figure}

Dialogue state tracking (DST) module, which aims to extract dialogue states during conversation \cite{DBLP:journals/pieee/YoungGTW13}, is an important component for task-oriented dialog systems to understand users' goals and needs \cite{DBLP:conf/eacl/Rojas-BarahonaG17,lei2018sequicity}. Dialogue states are sets of slots and their corresponding values as shown in Figure \ref{DST}. A \textit{slot} describes an attribute about the user's need (e.g. ``\textit{price range}'') and \textit{value} is the value of the given attribute (e.g. ``\textit{cheap}'' for ``\textit{price range}''). Collecting state labels can be costly \cite{DBLP:conf/emnlp/BudzianowskiWTC18}, requiring experts to annotate all \textit{(slot, value)} information for each turn in dialogues. In addition, dialogue states are various in different dialogue systems. For example, for different goods in an e-commerce platform, the types of users' needs are very different (e.g. ``size'' for clothes and ``CPU'' for computers). Thus, it is difficult to define all possible slots and their values in advance, especially with the wide application of dialogue systems in more and more new-rising applications. These challenges require DST models to be able to generate dialogue states in circumstances with limited annotations and knowledge about slot ontology.

To reduce the dependency on large amounts of training data, some few-shot methods are proposed recently for low-resource DST. Most of them apply domain transfer-based methods \cite{trade,DBLP:conf/aaai/LeeJ19,DBLP:conf/aaai/RastogiZSGK20} which rely on the assumption about the similarity among different domains and thus do not generalize very well to completely unseen slots.
Some approaches have tried to exploit external knowledge. \citet{DBLP:conf/asru/ChenWR13} and \citet{ddswws} consider slots and frames as similar semantic units and use the FrameNet semantic parsers to automatically induce slots. \citet{todbert} fine-tune BERT with a task-oriented dialogue dataset and utilize it for the downstream DST task. These methods rely on pre-defined slot ontology and can not generate unseen dialogue states. 


\begin{figure*}[t]
\centering
\includegraphics[width=1\textwidth]{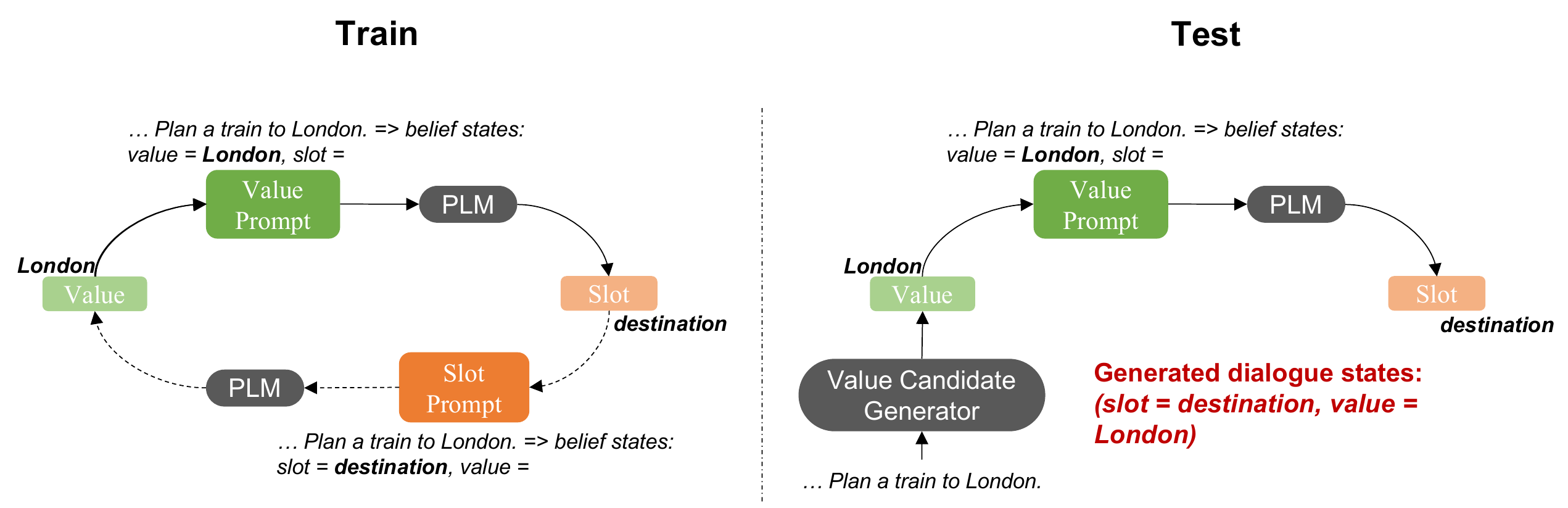}
\caption{Overview of the dual prompt learning framework for few-shot DST. While training, two dual prompts are constructed: {\color{DarkGreen}\textbf{value prompt}} and {\color{DarkOrange4}\textbf{slot prompt}}. Value prompt is constructed with a value and given to the PLM to generate corresponding slots. Slot prompt is constructed with slots and used to generate values. While testing, value candidates are first generated by a pre-trained value candidate generator, and then used to construct value prompts and generate slots.}
\label{train-arch}
\end{figure*}



In this paper, we rethink about DST task as a generation task. Considering slots and values as core semantic information that can be generated from dialogues, DST is similar to a hybrid summarization task including both extractive and abstractive summarization as target information can be both in and not in the original text. From the generation perspective, slots and values do not need to be predefined. We require a model which can understand such semantics and generate them as dialogue states. 
 
Recently, a new paradigm, “\textit{Pre-train, Prompt and Predict}” \cite{prompt-survey}, which aims at utilizing PLM in a more effective way, has aroused the public's attention. This paradigm can efficiently ``probe'' the target task-related knowledge with a textual \textit{prompt} and its superiority has been shown in many few-shot scenarios like few-shot text classification \cite{DBLP:conf/acl/GaoFC20} and text summarization \cite{DBLP:conf/acl/LiL20}. In 2021, \citet{pptod} use ``\textit{translate dialogue to belief state:}'' as prompts to generate dialogue state sequences. Such a simple prompt incorporates limited knowledge about DST task.
Thus, the promising paradigm is still very under-explored in low-resource DST task. To further exploit the potential of prompt learning, we design a dual prompt learning framework (DPL) for few-shot DST. Different from existing work which generates both slot and value in a sequence, we consider the learning of slot generation and value generation as dual tasks, and two prompts are designed based on such a dual structure to incorporate task-related knowledge of these two tasks respectively. In this way, DST task can be formulated as a language modeling task efficiently under few-shot settings. 

As shown in Figure \ref{train-arch}, we first design value prompt functions which equip the textual prompt with values and dialogue history. A value prompt function is a textual template, e.g., ``[$c$] belief states: value = [$v$], slot = [$s$]''. Given the dialogue history $c$ (``\textit{...Plan a train to London}'') and value candidate $v$ (\textit{London}), the prompt becomes: ``\textit{...Plan a train to London. belief states: value = London, slot = [s]}'' where \textit{[s]} is supposed to be generated as \textit{destination} by the PLM. 
Further, prompting values via slots can be seen as a dual task of prompting slots with values. Thus, we design slot prompt mechanism as the dashed lines in Figure \ref{train-arch} show. While training, after generating slots $s$ via value prompt, slots are presented to the slot prompt function $I$. This process aims to generate the corresponding value $v'$ which is supposed to be close to the original input $v$. Naturally, there exists an internal correlation between these two types of prompt tasks and they can benefit each other, especially under the few shot settings. The slot prompt can also help self-checking and restrict the output of the value prompt: if a generated slot can be used to prompt the original value, the value belongs to the slot with a larger probability. Finally, a simple but effective ensemble method is used to leverage the complementary advantages of different prompt functions while testing.

The main contributions of our work can be summarized as: 
\begin{itemize}
    \item We reformulate DST as a language modeling task and propose to split it into two dual tasks: slot generation and value generation.
    \item A novel dual prompt learning framework is designed to help PLMs understand the essence of DST with few labels and utilize the generation ability of PLMs efficiently.
    \item Experimental results show that our model can generate unseen dialogue states and outperforms state-of-the-art few-shot approaches.
\end{itemize}


\section{Preliminary}
\subsection{Prompt Learning}
Prompt learning, which aims to utilize pre-trained language models more effectively with the help of \textit{prompt}, is a new NLP paradigm (``\textit{Pre-train, Prompt and Predict}'') proposed recently. Usually, the original task input $x$ is used to construct a prompt that can reformulate the original task into a language modeling task. Take the emotion classification task as an example, when recognizing the emotion of a social media post, \textit{“I missed the bus today”}, we may continue with a prompt \textit{“I felt so \_\_”}, and ask the PLM to fill the blank with an emotion-bearing word. With the appropriate prompts, PLM can be pushed to generate the task-related output directly.


Given the \textit{prompt function f} which maps original input x to the prompt, the goal is to learn:
\begin{equation}
P\left(y \mid  f(x)\right)
\end{equation}
where $y$ is the \textit{answer} to be generated/filled. In DST, $y$ can be a word in the dialogue state sequence. 


\subsection{Dialogue State Tracking}
We consider each conversation with $T$-turn utterances alternating between the user and system: $C=\left\{a_1, u_1, ...a_T, u_T\right\}$ where $u_t$ and $a_t$ represent the user's and system's utterance respectively. Given the dialog history $c_t$ (including current user utterance $u_t$ and the former utterances, $c_t = \left\{a_1, u_1, ..., a_t, u_t\right\}$), a DST model aims to extract the dialogue state (belief state) $B_t$ for $u_t$ which comprises multiple tuples of slots $s$ and their associated values $v$ ($B_t = \left\{(s_1, v_1),...(s_n, v_n)\right\}$). For example, given the dialog history $c_t$ (``\textit{...Plan a train to London on this Tuesday}''), DST model is supposed to generate belief states $B_t =\{$\textit{(destination, London), (day, this Tuesday)}$\}$. The goal is to learn probability distribution $P$ \cite{soloist} for $t$-th turn:
\begin{equation}
P\left(B_{t} \mid c_t\right)
\label{eq:goal}
\end{equation}

If $B_t$ is considered as a word sequence \cite{simple-tod}, DST is essentially a language modeling task. Large-scale pre-trained language models (PLM) show outstanding language modeling and generation ability. Following the existing paradigm (\textit{Pre-train and Fine-tune}), we need to fine-tune PLM with the task-related dataset. Fine-tuning with a few labeled dataset may lead to over-fitting. Thus, an effective way to help PLM understand DST task in their familiar way (language modeling) and utilize the generation ability is important, inspiring the exploration of prompt learning for few-shot DST.

\section{Method}

\subsection{Dual Prompt Learning for Few-shot Dialogue State Tracking}
To utilize the few-shot generation ability of PLMs via prompt learning, previous work \cite{DBLP:journals/tacl/JiangXAN20,DBLP:conf/eacl/SchickS21,template-ner} show that the design of prompt function $f$ is a key factor that influences the final performance. The main question is how to formulate the downstream task as a language modeling task and thus can utilize the generation ability of PLMs efficiently. A natural idea is to consider slots and values as same semantics, dialogue history can be used as the input and fed into GPT2 to generate the sequence of dialogue states directly \cite{LinTB21}. However, this method needs plenty of annotations as the learning process lacks knowledge about the target task. 

Some existing work use slots as prompts and generate values \cite{zero-dst-qa}. For example, given $c$ (``\textit{...plan a train to London.}'') and slot (\textit{destination}), the input of PLM becomes ``\textit{...Plan a train to London. Where is the destination the user wants to reach? [$v$]}'' where \textit{[$v$]} is supposed to be generated as \textit{``London}''. This method relies on the known ontology of slot type. For few-shot DST, the slots that appear in the few labeled datasets may not include all possible needs. In addition, defining all possible slots are difficult as the rapid application in different new-rising domains and user's continuous need. In the real-world application, the candidate set of $s$ may be unknown and changeable.

Actually, values and slots are both core semantic units in utterances that describe users' needs. Generating values with slots can be seen as a dual task of generating slots with values. Naturally, these two types of tasks are supposed to hold an intrinsic correlation and can benefit each other, especially in the few-shot settings. Thus, we split the DST task into two dual sub-tasks (slot generation and value generation) and propose a dual prompt learning framework (DPL) for few-shot DST as shown in Figure \ref{train-arch}. While training, few labeled data can help PLMs understand DST under the dual prompt framework.
Next, we'll describe the framework in detail. 




\subsubsection{Value Prompt}\label{value-slot}
Four intuitive templates for value prompts are shown in Table \ref{prompts}. Take the first template $f_1$ for an example (\textit{``[c] belief states: value = [$v$], slot = [$s$]''}), given value candidate $v$ = \textit{``London''}, $f(c, v)$ = \textit{``[$c$] belief states: value = London, slot = [$s$]''}. The goal is to learn the probability of slots given $c$ and the value $v$:
\begin{equation}
P\left(s \mid  f(c, v)\right)
\end{equation}
The overall learning objective of this generation processing is minimizing the negative log-likelihood of slots in the training dataset $D$:
\begin{equation}
\mathcal L_v = -\sum_{i}^{|D|} log\ P\left(s_i \mid f(c_i, v_i)\right)
\label{eq:loss_v}
\end{equation}


As a turn may contain multiple values and slots, each pair of (slot, value) constructs an instance for training and testing. It's worth mentioning that the slot types are not static. We generate the slots in the whole vocabulary space, making generating unseen slots possible.


\begin{table}
\centering
\begin{tabular}{cl}
\hline
\multicolumn{2}{c}{\textbf{Prompt Functions}}\\
\hline
$f_1$ & \textit{[$c$] belief states: value = [$v$], slot = \textbf{[$s$]}} \\
$f_2$ & \textit{[$c$] belief states: [$v$] = \textbf{[$s$]}} \\
$f_3$ & \textit{[$c$] [$v$] is the value of \textbf{[$s$]}} \\
$f_4$ & \textit{[$c$] What is the slot type of [$v$] ? \textbf{[$s$]}} \\
\hline
\end{tabular}
\caption{\label{prompts}
Different value prompt functions. [$c$] is the dialogue history. [$v$] is the input of value candidate and [$s$] is the slot to be generated.
}
\end{table}




\subsubsection{Slot Prompt}\label{slot-value}


Although slots and values are all core semantics of dialogue, they are differently expressed. Slots are types, which are more often implicitly indicated in the dialogue. However, values are the specific needs users express. So they have more probability of explicitly appearing in the dialogue. We analyze the multi-domain dialogue dataset MultiWOZ 2.1 \citet{multiwoz2.1} and find that $89.36\%$ values can be matched in the original dialogue. So, slot prompt is considered as an auxiliary task for the value prompt. Our goal is to utilize it to help PLM understand the task and tune the output further: if a slot can be used to prompt the original value, it means there is a larger probability that the value belongs to the generated slot. 

Thus, we design \textit{slot prompt} as shown in Figure \ref{train-arch}. While training, the slot ($s$) is presented to slot prompt function ($I$). The slot prompt process aims to answer the corresponding value $v'$ which is supposed to be close to the original input one $v$. We take ``\textit{[$c$] belief states: [s] = [v]}'' as the template in $I$. We use teacher forcing for training and the loss function $\mathcal{L}_s$ is:
\begin{equation}
\mathcal{L}_s = -\sum_{i}^{|D|} log\ P\left(v_i \mid I(c_i, s_i)\right)
\label{eq:loss_s}
\end{equation}



\begin{algorithm}[tb]
    \caption{Training process}
    \label{alg:train}
    \begin{algorithmic}[1] 
        \REQUIRE Training dataset $D$, value prompt function $f$, slot prompt function $I$ and a PLM $g$\\
	\ENSURE The fine-tuned model $g$
        \REPEAT
        \FOR{batch $d \in D$}
        \STATE $d^v \leftarrow$ Initialize training set of value prompts as $\emptyset$
        \STATE $d^s \leftarrow$ Initialize training set of slot prompts as $\emptyset$
        \FOR{$i=1,...,|d|$}
        \STATE \textit{\#Get value prompts}
        \STATE $x_i^v \leftarrow f(c_i, v_i)$
        \STATE $d^v\ \leftarrow d^v\cup (x_i^v, s^i)$
        \STATE \textit{\#Get slot prompts}
        \STATE $x_i^s \leftarrow I(c_i, s_i)$
        \STATE $d^s\ \leftarrow d^s\cup (x_i^s, v^i)$
        \ENDFOR
        \STATE Calculate $\mathcal{L}_v$ on $d^v$ in Eq. \ref{eq:loss_v}
        \STATE Calculate $\mathcal{L}_s$ on $d^s$ in Eq. \ref{eq:loss_s}
        \STATE Calculate final loss $\mathcal L$ in Eq. \ref{eq:loss_final}
        \STATE Update $g$ to minimize $\mathcal L$
        \ENDFOR
        \UNTIL the training process converges
        \RETURN Updated $g$
	\end{algorithmic}
\end{algorithm}

\subsubsection{Training Process}
The final loss function $\mathcal L$ consists of loss functions in slot generation $\mathcal L_v$ and value generation $\mathcal{L}_s$:
\begin{equation}
\mathcal L = \mathcal L_v + w * \mathcal{L}_s
\label{eq:loss_final}
\end{equation}
where $w$ is a decimal in $(0,1)$ and is used to adjust these two tasks. The training process is described in Algorithm \ref{alg:train}.

\subsection{Inference}
\subsubsection{Valule Candidate Generation}
At training time, the labels of values are annotated and used for training. While testing, they are unknown. Existing work \cite{latent} extracts adjectives, named entities and others as value candidates. However, many values are implicit or do not belong to these pre-defined types. We consider the values in a turn as a sequence (e.g., $``Monday\ |\ 17:00"$) and the generation of values as a few-shot summarization task. 

As shown in Figure \ref{fig:vg}, the input is the dialog history $c$ concatenated with a prompt $``=> value :"$ and output is the value sequence ``$v_1\ |\ v_2\ |\ v_3\ |\ ...$''. The training dataset is the same as that used in the training process of DPL and loss function is also the negative log-likelihood loss $\mathcal L'$. Further, the trained model $g$ is utilized to tune the training process of value candidate generation towards generating values which can be used to generate correct slots via self-critical sequence training (SCST) \cite{rennie2017self}. After generating values, they are constructed as value prompt and fed into $g$ to generate slots. The loss of the tuning process is:

\begin{figure}[t]
\centering
\includegraphics[width=0.35\textwidth]{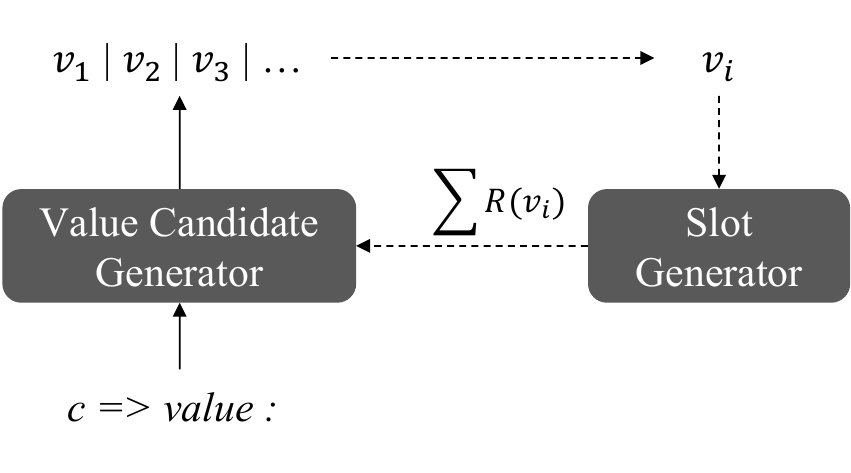}
\caption{The process of value candidate generation. Dashed lines denote the tuning process: The slot generator accepts the generated value $v_i$ to construct the value prompt and then feeds it into the fine-tuned PLM to generate slots and get reward for tuning the value candidate generator.}
\label{fig:vg}
\end{figure}

\begin{equation}
\mathcal L_r = - \sum_{i}^{|D|} R(v_i) * log\ P\left(v_i\right)
\label{eq:loss_r}
\end{equation}
where $v_t$ is the generated values and $R(v_i)$ is the reward which is the generation probability of target slot using $v_i$ as the input for the slot generator. 
The final loss is:
\begin{equation}
\mathcal L'' = \lambda \mathcal L' + (1-\lambda) \mathcal L_r
\label{eq:loss_vg}
\end{equation}
where $\lambda$ is a decimal in $(0,1)$ and is used to balance these two losses.


\subsubsection{Prompt Ensemble}\label{prompt ensemble}
In Section \ref{value-slot}, we described methods to generate a set of value prompt functions as shown in Table \ref{prompts}. Each of these prompts may be more or less effective at eliciting knowledge from PLMs, and thus it is necessary to decide how to use these generated prompts at test time. Unfortunately, under few-shot settings, it's hard to get enough training and development set to automatically select or generate the best-performing prompt \cite{DBLP:journals/tacl/JiangXAN20,DBLP:conf/acl/GaoFC20,ben2021pada,DBLP:conf/emnlp/DavisonFR19,DBLP:journals/corr/abs-2103-10385}. We introduce a multi-prompt learning method (\textit{prompt ensemble}) for few-shot DST task in this section to effectively utilize different prompts.


Prompt ensemble methods use multiple unanswered prompts for input at inference time to make predictions \cite{prompt-survey}. It can leverage the complementary advantages of different prompts and alleviate the cost of choosing one best-performing prompt. There is relatively little work on prompt ensemble for generation tasks. A simple way for ensemble in this case is to train a separate model for each prompt and generate the output based on the vocabulary distribution learned by several models while testing. The probability of slot $s_i$ is calculated via:


\begin{equation}
P\left(s_i \right)=  \sum_{k}^{K} \alpha_k * P\left(s_i \mid f_k(c_i,v_i)\right)
\label{eq:ensemble}
\end{equation}
where $f_k$ is the $k$-th prompt and $\alpha_k$ is its weight. $K$ is the number of prompt functions.

\section{Experiments}
\subsection{Experimental Setup}

\subsubsection{Dataset and Evaluation}
We evaluate our methods on MultiWOZ 2.0 \cite{DBLP:conf/emnlp/BudzianowskiWTC18} and 2.1 \cite{multiwoz2.1} dataset. MultiWOZ 2.1 is a multi-domain task-oriented dialog dataset with 7 domains and fixes some noisy state annotations in the MultiWOZ 2.0. Following existing work~\cite{trade}, we keep five domains (\textit{Attraction, Hotel, Restaurant, Taxi, Train}) because the other two domains only appear in the training set. Each turn can include multiple slots. 


We adopt the standard metric in DST \cite{trade}: joint goal accuracy (JGA). The metric compares the entire predicted belief states to the gold one at each dialog turn. The prediction is considered correct if and only if all the predicted states exactly match the ground truth states. Only when the values and slots are both correct, the prediction is correct.

\subsubsection{Implement Details}
We choose SOLOIST \cite{soloist} as our base model. SOLOIST is initialized with the 12-layer GPT-2 \cite{gpt2} and further trained on multiple task-oriented dialog corpora (Schema \cite{sgd} and Taskmaster \cite{taskmaster} ) for two dialogue-related tasks (belief prediction and response generation). Specifically, the belief prediction task accepts utterance as input and generates the belief states as a word sequence (e.g., ``\textit{Belief state: destination = London}''). Thus, we suppose that knowledge about DST may be learned via SOLOIST, and what we need to do is to find an effective way to ``probe'' the knowledge and apply it to few-shot scenarios. In addition, the moderate size of SOLOIST (117M parameters) makes fine-tuning for the task-related prompts computationally efficient. $\alpha$ for each prompt function in Eq.\ref{eq:ensemble} is set to the same value (1/4). $w$ in Eq.\ref{eq:loss_final} is 0.1. $\lambda$ in Eq. \ref{eq:loss_vg} is set to 0.1. Our code will be released after the review process.



\subsection{Few-shot Experiments}
For few-shot experiments, we compare our methods with several strong baselines capable of few-shot inference, which achieve SoTA on MultiWOZ 2.0 dataset. They can be categorized into two classes: one need slot ontology and another doesn't need it. Baselines requiring slot ontology include: (1) \textbf{TRADE} \cite{trade} requires the embedding of slots as inputs and uses a soft copy mechanism to either copy the corresponding values from utterance pairs or generate them using RNN. (2) \textbf{Self-Sup} \cite{self-sup} adds two self-supervised objectives: preserving latent consistency and modeling conversational behavior for TRADE. (3) \textbf{TOD-BERT} \cite{todbert} trained BERT with several task-oriented dialogue-relevant tasks: masked language modeling and response generation with large-scale corpora (100k dialogues across over 60 different domains). For DST, it learns a classifier to predict the value over the pre-defined possible value set for each known slot.
Baselines that do not needs slot ontology consider DST as sequence generation task: (1) \textbf{SimpleTOD} \cite{simple-tod} uses a single causal language model to generate all outputs given the dialogue context (2) \textbf{MinTL} \cite{mintl} jointly learn DST and dialogue response generation which introduces Levenshtein belief spans. (3) \textbf{SOLOIST} \cite{soloist} is the base model of DPL. (4) \textbf{PPTOD} \cite{pptod} integrates different dialogue modules into a unified model with prompt.

To simulate the few-shot scenarios, we randomly select a limited quantity of labeled training data for training. To compare with previous work, we randomly select dataset with given ratio (1\%, 5\%, 10\%, 20\% and 25\%) in the training set for training and test on the whole test set in Table \ref{fewshot}. Some baselines provide results of 20\% training set while others provide that of 25\%. For comparison, we evaluate our model on both 20\% and 25\% settings. ``N/A'' denotes the results not presented in the original paper. 


Compared to previous approaches, our model achieves consistently higher
JGA (3.9\% on average) on other domains under different data ratio settings. The improvement is especially large when only 1\% training set are available (4.0\% over the strong baseline PPTOD). It indicates the superiority of our model in low-resource scenarios and verifies the strong task-related generation ability of PLMs under prompt learning.

\begin{table}[t]
\centering
\begin{tabular}{l|c|c|c|c|c}
\hline
& 1\% & 5\% & 10\% & 20\% & 25\% \\
\hline
\multicolumn{6}{c}{\textit{Need slot ontology}} \\
\hline
TRADE & 9.7 & 29.4 & 34.1 & N/A & 41.4 \\
Self-Sup & 20.4 & 33.7 & 37.2 & N/A & 42.7 \\
TOD-BERT & 10.3 & 27.8 & 38.8 & N/A & 44.3 \\
\hline
\multicolumn{6}{c}{\textit{No need for slot ontology}} \\
\hline
SimpleTOD & 7.9 & 16.1 & 22.4 & 31.2 & N/A\\
MinTL & 9.3 & 21.3 & 30.3 & 36.0 & N/A\\
SOLOIST & 13.2 & 26.5 & 32.4 & 38.7 & N/A\\
PPTOD & 29.7 & 40.2 & 43.5 & 47.0 & N/A\\
DPL & \textbf{33.7} & \textbf{42.1} & \textbf{45.6} & \textbf{49.5} & \textbf{51.2}\\
\hline
\end{tabular}
\caption{\label{fewshot}
Few-shot experimental results on MultiWOZ 2.0. 
}
\end{table}

\begin{table}
\centering
\begin{tabular}{p{0.95\columnwidth}}
\hline
slots\\
\hline
$area^{123}$, $arrive\ by^{45}$, $day^{235}$, $departure^{45}$, $destination^{45}$, 
$food^3$, $internet^2$, $leave^{45}$, $name^{123}$, $people^{235}$,
$parking^2$, $price^{23}$, $stars^2$, $stay^2$, $time^3$, $type^{12}$\\
\hline
\end{tabular}
\caption{\label{slots}
All slots in MultiWOZ 2.1. The upper script on slot indicates the domain it belongs to (1: \textit{Attraction}, 2: \textit{Hotel}, 3: \textit{Restaurant}, 4: \textit{Taxi}, 5: \textit{Train}).
}
\end{table} 

\subsection{Few-shot Cross-domain Experiments}
In the few-shot cross-domain experiments, models are first trained with four domains and then fine-tuned with 1\%, 5\% and 10\% of the target domain data. We compare with several strong models with reported results: \textbf{TRADE}, \textbf{DSTQA} and \textbf{T5DST} \cite{T5DST}. DSTQA considers DST as question-answer task and needs slot information to construct questions. T5DST is a strong prompt baseline that uses slot descriptions as a prompt. They all rely on known slot ontology. The experiments are also conducted on MultiWOZ 2.0 for comparison with previous works. Table \ref{cross-domain-fewshot} summarizes the evaluation results. We can see that in all domains, our model outperforms these strong baselines, especially in 1\% training data setting for hotel domain. 

\begin{table*}[t]
\centering
\begin{tabular}{lccccccccccccccc}
\hline
\multirow{2}*{Model} & \multicolumn{3}{c}{Attraction} & \multicolumn{3}{c}{Hotel} & \multicolumn{3}{c}{Restaurant} & \multicolumn{3}{c}{Taxi} & \multicolumn{3}{c}{Train} \\
 & 1\% & 5\% & 10\% & 1\% & 5\% & 10\% & 1\% & 5\% & 10\% & 1\% & 5\% & 10\% & 1\% & 5\% & 10\%\\
\hline
TRADE & 35.8 & 57.5 & 63.1 & 19.7 & 37.4 & 41.4 & 42.4 & 55.7 & 60.9 & 63.8 & 66.5 & 70.1 & 59.8 & 69.2 & 71.1 \\
DSTQA & N/A & 70.5 & 71.6 & N/A & 50.2 & 53.7 & N/A & 59.0 & 64.5 & N/A & 70.9 & 74.2 & N/A & 70.4 & 74.5 \\
T5DST & 58.8 & 65.7 & 69.5 & 43.1 & 50.7 & 54.9 & 57.6 & 61.9 & 63.5 & 70.1 & 73.7 & 74.7 & 70.8 & 74.2 & 77.6 \\
\hline
DPL & \textbf{60.4} & \textbf{70.5} & \textbf{72.1} & \textbf{45.7} & \textbf{53.1} & \textbf{56.9} & \textbf{60.5} & \textbf{64.3} & \textbf{67.2} & \textbf{74.1} & \textbf{76.4} & \textbf{77.8} & \textbf{72.1} & \textbf{76.3} & \textbf{79.0} \\
\hline
\end{tabular}
\caption{\label{cross-domain-fewshot}
Few-shot cross-domain experimental results on MultiWOZ 2.0.
}
\end{table*}


\subsection{Unseen Slot Generation}
We present the slots of each domain in Table \ref{slots}. We find that some domains share some slots with other domains. For example, all slots of \textit{Attraction} can be found in \textit{Hotel}. On the contrary, some domains hold some slots that are not seen in other domains. For \textit{Hotel}, it has four unseen slots: \textit{parking}, \textit{book stay}, \textit{stars} and \textit{internet}. \textit{Restaurant} has two unseen slots (\textit{food} and \textit{time}). Here, we consider ``unseen slots'' as both ``unseen'' in the labeled training data and ``unseen'' in the slot names of the source domains.

\begin{figure}[t]
\centering
\includegraphics[width=0.5\textwidth]{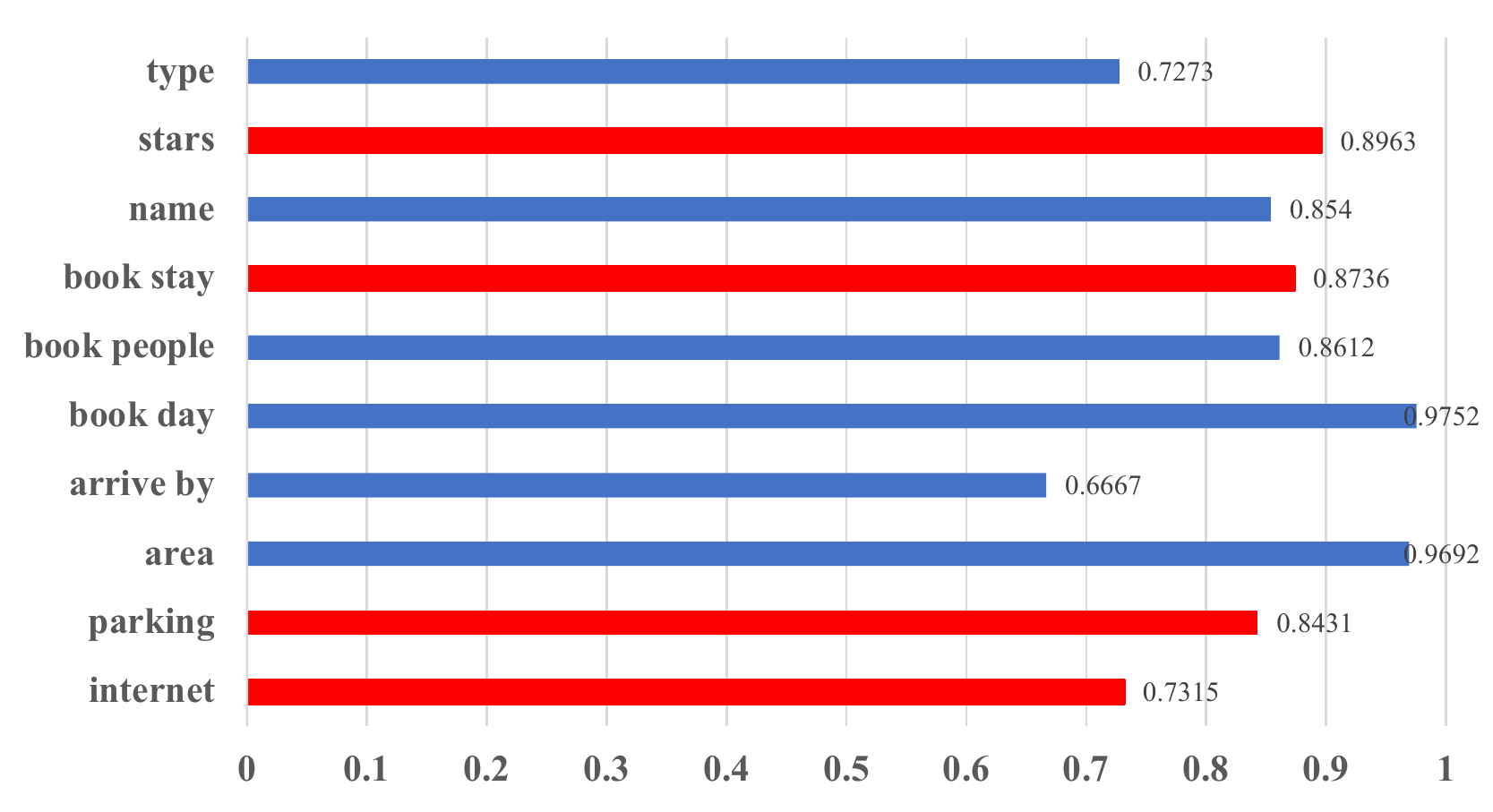}
\caption{Slot accuracy of each slot in \textbf{Hotel} domain under zero-shot settings. X-axis is the slot accuracy and y-axis is the slot. Red bars mark {\color{red}unseen slots}.}
\label{hotel}
\end{figure}

\begin{figure}[t]
\centering
\includegraphics[width=0.5\textwidth]{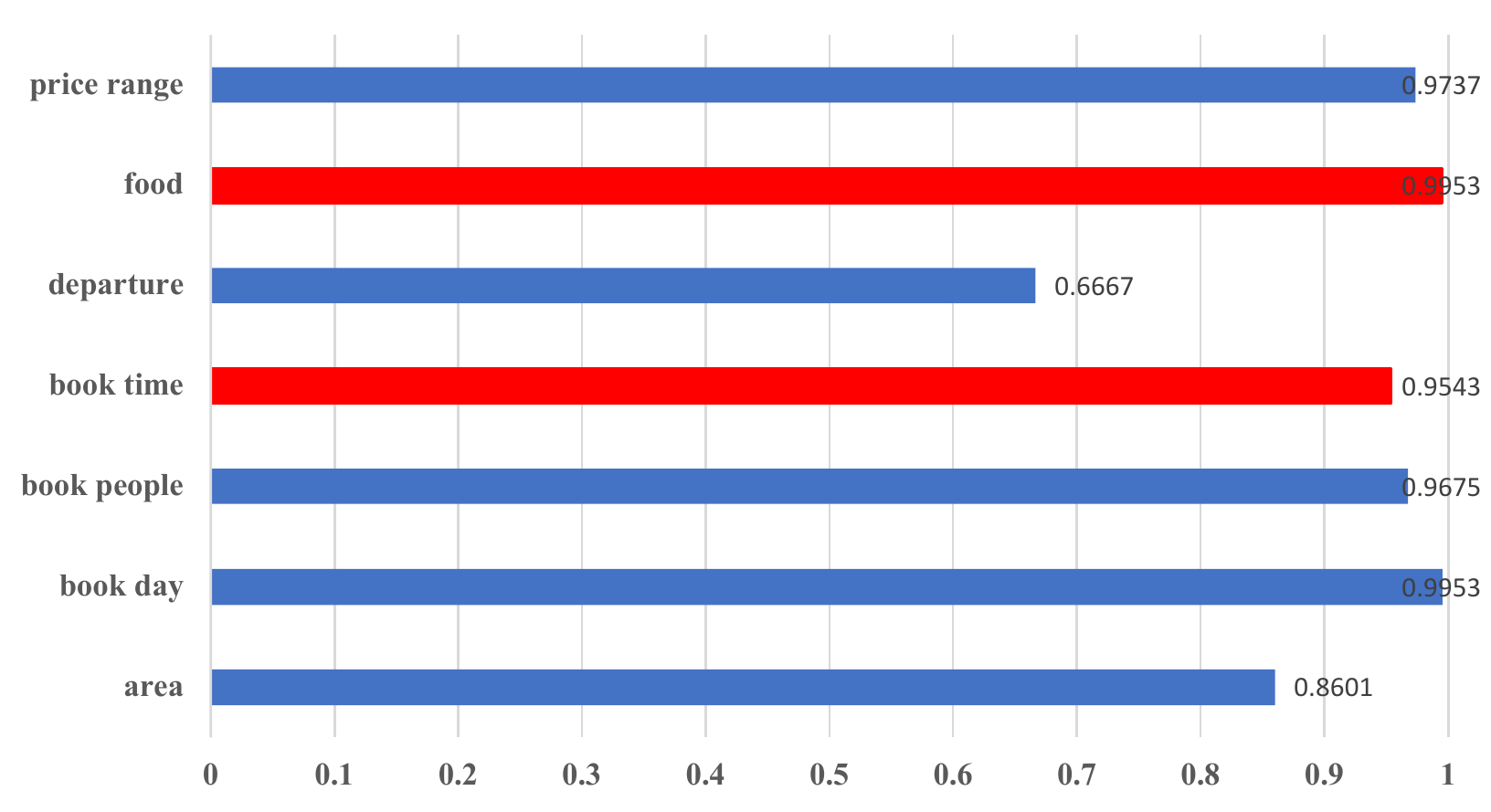}
\caption{Slot accuracy of each slot in \textbf{Restaurant}.}
\label{restaurant}
\end{figure}

To observe the extension and generation ability for unseen slots, we design two zero-shot experiments: leave \textit{Hotel} or \textit{Restaurant} as held-out-domain respectively, and train on other four domains. We present slots accuracy which evaluates the slot-level accuracy of correctly generated slots while values are correctly generated. From the results in Figure \ref{hotel} and \ref{restaurant}, we find that: 

(1) For seen slots that have the same names as that of source domains, our model can generate them with high accuracy. For example, \textit{area} in \textit{Hotel} domain is a common slot for other two source domains (\textit{Attraction} and \textit{Restaurant}), which can be generated with 96.92\% accuracy. It indicates good transfer ability across domains.

(2) For some unseen slots (\textit{book stay} and \textit{stars} in \textit{Hotel} of Figure \ref{hotel}, \textit{book time} and \textit{food} in \textit{Restaurant} of Figure \ref{restaurant}), our model can generate them with more than 87\% accuracy. 
For example, given the dialogue history ``\textit{...yes, please book it for 1 person and for 5 nights starting Friday.}'' The model successfully generates ``\textit{book stay}'' for ``\textit{5}'' even it has never seen the instances of book stay while training. Without known slot types, our model can infer the hidden semantic from the value and contexts, which is supposed to be the slot.

(3) For two unseen slots (\textit{internet} and \textit{parking}), their values are ``yes''. We find that the value generator can generate such implicit value as shown in Table \ref{a-example}. Then PLM model can generate the corresponding slot with large probabilities (73.15\% and 84.31\%).

\subsection{Ablation Studies}

\subsubsection{Value Candidate Generation}
We then analyze the results of value generation given the corresponding ratio of training data. Table \ref{value_generator} presents turn-level accuracy which measures the ratio of turns while all predicted values exactly match the ground truth values.

Rule-based candidate generator achieves $32.65\%$ turn-level accuracy. Our trained generator can outperform it with only $1\%$ training data over $14\%$, indicating the superiority of learned value generator. In addition, we find that ``tuning'' can improve the results of value generation. Although the value generator didn't achieve very high values of turn-level accuracy under the few-shot setting,
our model still outperforms others as JGA in Table \ref{fewshot} shows. It attributes to the high accuracy of slot generation while turn-level values are correctly generated. 
%

\begin{table}[t]
\centering
\begin{tabular}{l|c|c|c|c}
\hline
rule & \multicolumn{4}{c}{32.65}\\
\hline
& 1\% & 5\% & 10\% & 25\% \\
\hline
Ours & 51.42 & 59.22 & 63.11 & 65.17\\
Ours \textit{w/o tuning} & 47.58 & 55.93 & 61.57 & 65.03 \\
\hline
\end{tabular}
\caption{\label{value_generator}
Turn-level accuracy on test set of value generator under different ratios of training data. ``\textit{w/o tuning}'' means removing the process of using the output of slot generation to tune the process of value generation. 
}
\end{table}

\begin{table}[t]
\centering
\begin{tabular}{p{0.95\columnwidth}}
\hline
\textbf{Dialogue history:} ... [user] no , i do not care where it is . i like 3 stars and i absolutely need free wifi . \\
\textbf{Gold values: } don't care, 3, yes \\
\textbf{Generated values: } don't care, 3, yes \\
\hline
\end{tabular}
\caption{\label{a-example}
 A test instance whose values are generated by the trained value generator with 25\% training data. It shows that the value generator can generate implicit values (``yes'').
}
\end{table}

\subsubsection{Prompt Functions}

We further observe the performances of different components including different value prompt functions and prompt ensemble. We train separate models with each value prompt (``DPL'' for $f_1, ...f_4$). Then, we apply prompt ensemble (``En'') for the trained models. Experiments with 1\% training data are shown in Table \ref{ablstudy}. 
\begin{itemize}
    \item The first four numerals in the first row show the original performance with different prompt functions. Among the four prompts, $f_2$ performs best which may attribute to the similar format of $f_2$ compared with the output sequences in a pre-training task of SOLOIST (Considering dialogue history as inputs and generate dialogue states in the format as ``\textit{belief states: [s1] = [v1], [s2] = [v2]}'').
    \item The prompt ensemble enables further improvement. Under few-shot settings, prompt ensemble is a simple but efficient way of utilizing different prompt functions.
\end{itemize}

\subsubsection{Dual Framework}
In our dual framework, if we remove the branch of slot prompt (value generation), the model also can learn to generate slots based on value prompt. So we remove slot prompt to see its effects on the entire framework. Experimental results are reported in the ``DPL \textit{w/o slot prompt}'' row of Table \ref{ablstudy}. We find that the performances decrease for all prompt functions, indicating the importance of using slot prompt. For $f_2$, the decrease is relatively small (0.3\%). It may attribute to the slot prompt (``\textit{belief states: [s] = [v]}'') and the value prompt (``\textit{belief states: [v] = [s]}'') are too similar to learn complementary knowledge.


\begin{table}[t]
\centering
\begin{tabular}{l|c|c|c|c||c}
\hline
& $f_1$ & $f_2$ & $f_3$ & $f_4$ & $En$ \\
\hline
DPL & 25.7 & 29.4 & 26.4 & 28.9 & 33.7 \\ 
DPL \textit{w/o slot prompt} & 20.1 & 29.1 & 22.3 & 24.5 & 29.5 \\
\hline
\end{tabular}
\caption{\label{ablstudy}
JGA results for our models trained with 1\% data given different prompt functions (from $f_1$ to $f_4$). ``\textit{w/o slot prompt}'' means removing the training process of slot prompt. ``En'' shows the result of the ensemble of models trained on different prompt functions with and without slot prompt.
}
\end{table}

\begin{figure}[t]
\centering
\includegraphics[width=0.5\textwidth]{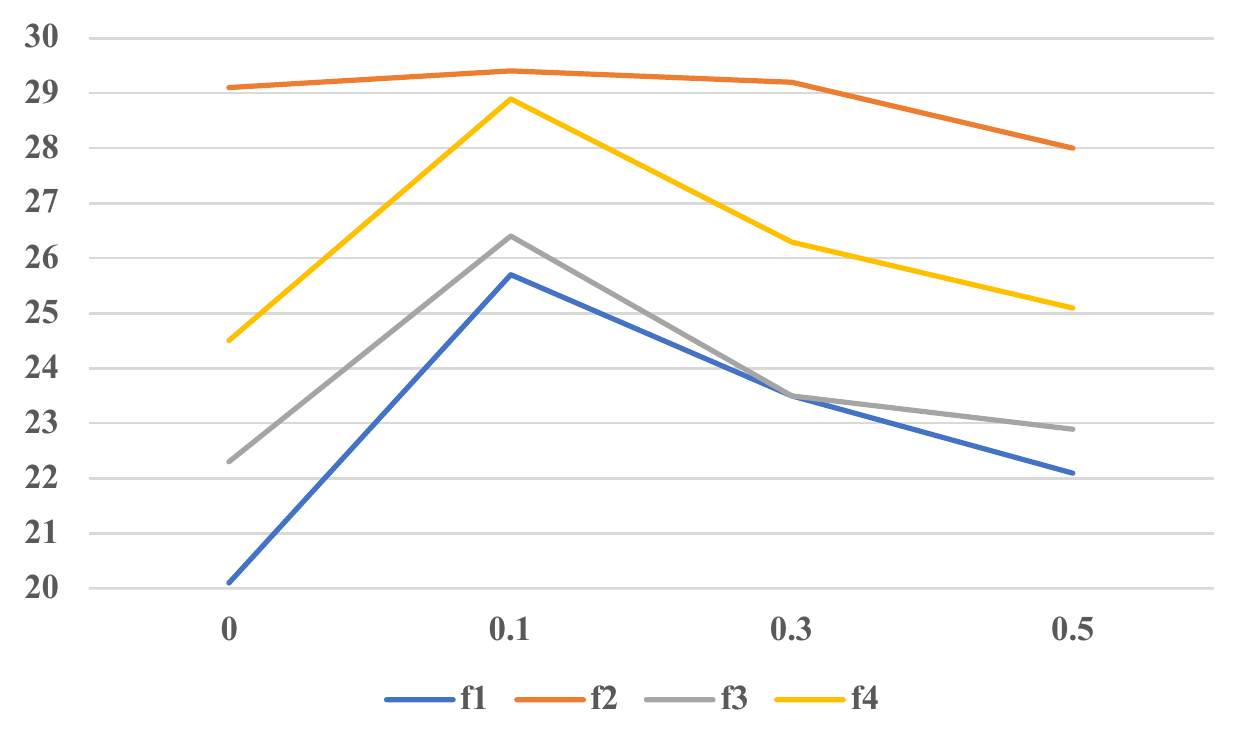}
\caption{The influence of weight $w$ for slot prompt using different prompt functions $f$. X-axis is the value of $w$ and y-axis is JGA. Experiments with $w=0.1$ always perform best for all prompt functions.}
\label{weight}
\end{figure}

Further, we conduct experiments to observe the influence of weight $w$ in Eq.\ref{eq:loss_final}. $w$ is set to $\{0, 0.1, 0.3, 0.5\}$. Experiments using 1\% training data and different value prompts are shown in Figure \ref{weight}. We find that the JGA performance always increases with the value of $w$ first and then begins to decrease. It means that slot prompt is actually an auxiliary task and can provide useful knowledge when the weight is relatively small. All experiments for the four prompts perform best when the $w$ is 0.1. So we set it to 0.1 in all experiments. 

\section{Related Work}

\subsection{Few-Shot Dialogue State Tracking}

Some few-shot methods used data augmentation to get more labeled data for training. \citet{DBLP:conf/acl/CampagnaFML20} and \citet{HouCCCL21} propose to synthesize dialogues for a new domain using the small number of domain templates derived from observing a small dataset and the ontology of the domain. These methods depend on the ontology of slots on the target domain.

Most of the existing work focuses on transferring from other resource-rich DST domains. \citet{DBLP:conf/aaai/LeeJ19} and \citet{DBLP:conf/aaai/RastogiZSGK20} utilize the slot description for transferring reusable concepts across domains. \citet{todbert} learn similarity functions between slots and values, and transfer them into unseen domains. \citet{d-reptile} introduces meta-learning and uses source domains to meta-learn the parameters of the model used to initialize the fine-tuning process of the target domain. One constraint of such methods is that they rely on domain similarity for transfer, and therefore cannot be applied to general domains.

Another thread of approaches tries to exploit external knowledge. \citet{DBLP:conf/asru/ChenWR13} and \citet{ddswws} utilize FrameNet-style \cite{framenet} semantic frames and named entity recognition (NER) as the weak supervision for slot candidates. \citet{rc2},\citet{rc1},  \citet{zero-dst-qa} and \citet{transferQA} reformulate DST into a Reading Comprehension (RC) task and make use of the abundant RC data and frameworks to overcome the data scarcity issue in the DST task. \citet{self-sup} investigate two self-supervised objectives: preserving latent consistency and modeling conversational behavior. However, they have limited performance owing to the limited common knowledge.



\subsection{Prompt Learning}
With the rapid development of large-scale pre-trained language models (PLM), a new paradigm arise public's attention: ``\textit{pre-train, prompt, and predict} \cite{prompt-survey}''. Instead of adapting PLM to downstream tasks via objective engineering, prompt learning reformulates downstream tasks to look more like those solved during the original PLM training with the help of a textual prompt. GPT-3 model \cite{gpt3} achieves remarkable few-shot performance solely by leveraging a few task demonstrations as input context (e.g., \textit{``Translate English into French''}) and a natural-language prompt (e.g., \textit{``cheese ==> ''}). However, training such a huge model (175B parameters) is difficult. A more usual prompt learning method is ``prompt-based fine-tune'': utilize a moderately-sized PLM for which fine-tuning is computationally efficient and fine-tune it with the task-related prompts. It shows good performance in many few-shot scenarios. \citet{DBLP:conf/acl/GaoFC20} use RoBERT-large and design automatic prompt generation for text classification. \citet{DBLP:conf/acl/LiL20} add continuous task-specific vector as prompt to each transformer layer and achieve improvements in low-resource text summarization. For DST task, \citet{sup-dst-prompt} use slots as prompt directly and generate the corresponding values, which needs a lot of labeled training data for fine-tuning PLM. For few-shot DST, the prompt learning-based methods are still under-explored.


\section{Conclusion}
For the lack of labeled data in practical DST tasks, we design a dual prompt learning framework, which consists of two main components (value prompt and slot prompt). Our model can effectively probe DST-related knowledge from pre-trained language models and utilize it for DST task. Experiments show that our model outperforms existing state-of-the-art methods under different levels of resources. In addition, this framework doesn't rely on the known ontology of slot types. With extensive experiments, we find that it can generate slots that are not seen in source domains and are not pre-defined as well with high probabilities. In the future, we'll focus on improving the performance of extracting value candidates.

\bibliography{sample-base}
\bibliographystyle{ACM-Reference-Format}
\end{document}